# Multi-task learning from fixed-wing UAV images for 2D/3D city modeling

M. R. Bayanlou[1, *,] M. Khoshboresh-Masouleh[2]

[1] Aerospace Engineering Department, Sharif University of Technology, Tehran, Iran - mohammadreza.bayanlou@ae.sharif.edu
[2] School of Surveying and Geospatial Eng., College of Eng., University of Tehran, Tehran, Iran - m.khoshboresh@ut.ac.ir

**ABSTRACT:**

Single-task learning in artificial neural networks will be able to learn the model very well, and the benefits brought by transferring knowledge thus become limited. In this regard, when the number of tasks increases (e.g., semantic segmentation, panoptic segmentation, monocular depth estimation, and 3D point cloud), duplicate information may exist across tasks, and the improvement becomes less significant. Multi-task learning has emerged as a solution to knowledge-transfer issues and is an approach to scene understanding which involves multiple related tasks each with potentially limited training data. Multi-task learning improves generalization by leveraging the domain-specific information contained in the training data of related tasks. In urban management applications such as infrastructure development, traffic monitoring, smart 3D cities, and change detection, automated multi-task data analysis for scene understanding based on the semantic, instance, and panoptic annotation, as well as monocular depth estimation, is required to generate precise urban models. In this study, a common framework for the performance assessment of multi-task learning methods from fixed-wing UAV images for 2D/3D city modeling is presented.

## 1. INTRODUCTION

In recent years, the role of traditional methods such as terrestrial mapping and traditional aerial photogrammetry techniques has been dimmed due to the high cost and also the need for a long time to generate a multi-task dataset for scene understanding (Masouleh and Sadeghian, 2019; Zhang and Yang, 2018). An affordable and accurate way to generate multi-task data is to use the combination of an Unmanned Aerial Vehicle (UAV) with a high-resolution digital camera (e.g., RGB, Multi-spectral, Thermal, or Hyperspectral) and machine learning methods (Bayanlou and Khoshboresh-Masouleh, 2020; Khoshboresh-Masouleh and Hasanlou, 2020). Although UAV with a high-resolution digital camera is an efficient tool for data generation, there is still a lack of multi-task datasets for scene understanding (Khoshboresh-Masouleh and Akhoondzadeh, 2021). With this study, we aim at promoting research on multi-task learning by generating a very high-resolution low-cost dataset of urban and rural objects (e.g., building, parcel boundary, vehicle, building shadow, vegetation, ground, waste object, farmland, and water). In this study, we focus on multi-task learning based on semantic segmentation, building panoptic segmentation, and monocular depth estimation.

A major yet unsolved research topic for accurate 2D/3D city model generation is multi-task learning for scene understanding from high-resolution low-cost photogrammetry and remote sensing data sources (Khoshboresh Masouleh and Saradjian, 2019). In remote sensing and photogrammetry, previous benchmarks include four semantic segmentation datasets designed using satellite, airborne, and UAV platforms for urban scene analysis.

The ISPRS 2D semantic labeling benchmark provides Vaihingen and the Potsdam datasets targeting semantic labeling for the urban scenes. The Vaihingen and the Potsdam datasets are 9 cm and 5 cm resolutions, respectively. There are 6 classes defined for the semantic segmentation task, including impervious surfaces, buildings, low vegetation, tree, car, and background. The ISPRS 2D semantic labeling benchmarks include a set of homogenous scenes from one spatial location, and most deep learning-based methods achieve high accuracy using these kinds of datasets. In recent years, many models have achieved high accuracy (mostly above 90%) on these test data (hereafter called Dataset 1 for short).

The ISPRS UAV Semantic Video Segmentation benchmark provides a very high-resolution video dataset targeting semantic labeling for urban scene analysis from an oblique UAV perspective. There are 8 classes defined for this dataset, including building, road, tree, low vegetation, static car, moving car, human, and background (Lyu et al., 2020). This dataset consists of only RGB images and is not suitable for multi-task learning (hereafter called Dataset 2 for short).

The ISPRS Benchmark Challenge on Large Scale Classification of VHR Geospatial Data provides a multispectral high-resolution satellite dataset targeting semantic labeling for urban scene analysis from two Worldview-II satellite images. There are 6 classes defined for this dataset, including impervious surface, building, pervious surface, high vegetation, cars, and water. Moreover, buildings are annotated as single objects for semantic instance segmentation (Roscher et al., 2020). This dataset consists of only remote sensing optical images and is not suitable for multi-task learning (hereafter called Dataset 3 for short).

In contrast to these datasets, our proposed dataset will be comprised of much larger multi-task data (e.g., semantic annotation, building panoptic annotation, and depth) and with more scene complexity in terms of the number of objects (e.g., parcel boundary and building shadow), which makes our dataset more adequate for multi-task learning for scene understanding from UAV images. An overview of existing datasets of annotated imagery can be found in Table 1.

| **Reference** | Dataset 1 | Dataset 2 | Dataset 3 | Ours |
|---|---|---|---|---|
| **Data source** | Airborne | UAV | Satellite (WV-2) | UAV |

* Corresponding author

| Type | Orthophoto | Video | Nadir view | Orthophoto |
|---|---|---|---|---|
| DSM/nDSM | High-resolution | No | No | Very high-resolution |
| Texture distortion | High | - | - | Low (cf. Figure 3) |
| Semantic annotation | Yes | Yes | Yes | Yes |
| Semantic classes | 6 | 8 | 6 | 8-10 |
| Panoptic annotation | No | No | No (instance) | Yes (just buildings) |
| Color-based 3D point cloud | No | No | No | Yes |
| Image size (pix) | 6000×6000 | 4096×2160 | 3452×3504 | 2000×2000 |
| GSD | 9cm/5cm | - | 50cm | 2.5cm |
| Is this a multi-task dataset? | Yes (low potential), including orthophoto, nDSM, and semantic annotation | No | No | Yes (high potential), including orthophoto, DSM, semantic, and building panoptic annotation |
| Number of labeled multi-task pixels | 2014: 1.7 million 2015: 1.4 billion | - | - | 2 billion |

Table 1. List of the previous datasets and the proposed dataset for 2D/3D urban scene analysis

## 2. SINGLE-TASK VS. MULTI-TASK LEARNING

Single-task learning in deep neural networks (LeCun et al., 2015) will be able to learn the model very well, and the benefits brought by transferring knowledge thus become limited. In this regard, when the number of tasks increases (e.g., semantic segmentation, panoptic segmentation, and monocular depth estimation), duplicate information may exist across tasks, and the improvement becomes less significant (cf. Figure 1a). Multi-task learning has emerged as a solution to knowledge-transfer issues (Cipolla et al., 2018). Multi-task learning is an approach to scene understanding which involves multiple related tasks each with potentially limited training data (cf. Figure 1b). Multi-task learning improves generalization by leveraging the domain-specific information contained in the training data of related tasks. In urban management applications such as infrastructure development, traffic monitoring, smart 3D cities, and change detection, automated multi-task data analysis for scene understanding based on the semantic, instance, and panoptic annotation, as well as monocular depth estimation, is required to generate precise urban models (Khoshboresh-Masouleh et al., 2020).

## 3. DATA COLLECTION

SAMA-VTOL aerial image dataset is a new open UAV-based image dataset for a wide range of scientific projects in remote sensing (e.g., 3D object modeling, rural/urban mapping, and digital elevation/surface model processing). Open high-quality UAV images play an important part in providing and expanding spatial data processing methods (Herwitz et al., 2004; Ishida et al., 2018; Senthilnath et al., 2017). This dataset includes 120 rural/urban scene images with 80% overlap between images (forward overlap) and 60% overlap between flight lines (side overlap) from part of Esfahan province, Iran. The characteristics that make the proposed dataset an excellent scientific dataset are: (i) very high ground sampling distance (GSD) due to suitable fly height selection; (ii) GNSS-PPK (Post Processing Kinematic) system for improving the spatial accuracy without ground control points (GCPs); (iii) various landscape types (e.g., different types of roofs for commercial/residential buildings, and vegetation), and (iv) uses of the new UAV-photogrammetry platform, named SAMA-VTOL (2019) has been developed by TAREQH Corporation (Bayanlou and Khoshboresh-Masouleh, 2020).

### 3.1 UAV Images

The original UAV RGB images were captured by SAMA-VTOL are provided for the case study. SAMA-VTOL is designed to boost efficiency, safety, and quality for high-resolution low-cost data collection. This dataset consists of 120 rural/urban scene images with 80% forward overlap and 60% side overlap, where the proposed dataset uses the WGS 84 (EPSG::4326) coordinate system, as do most GNSS units. Figure 2 shows the study site in the various landscape types of images collected from Esfahan province. The research site is part of the Esfahan province, Iran. The land cover consists of agricultural land and urban areas.

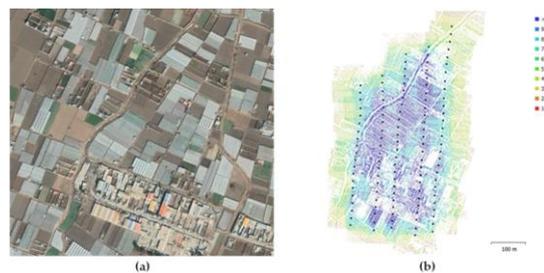

Figure 2. Google Earth imagery of the study area. (a) Research site; (b) Camera locations and image overlap

In this study, SAMA-VTOL was equipped with a Fujifilm X-A3 camera to acquire images. Additionally, the Agisoft Metashape software was used to analyzing images and produce dense point clouds, digital surface model (DSM), and orthoimage for evaluating quality and quantity, and QGroundControl software was used for mission planning and flight control.

**3.2 Data Processing**

The data processing includes automatic aerial triangulation-based bundle block adjustment with camera calibration and model generation by Agisoft Metashape.

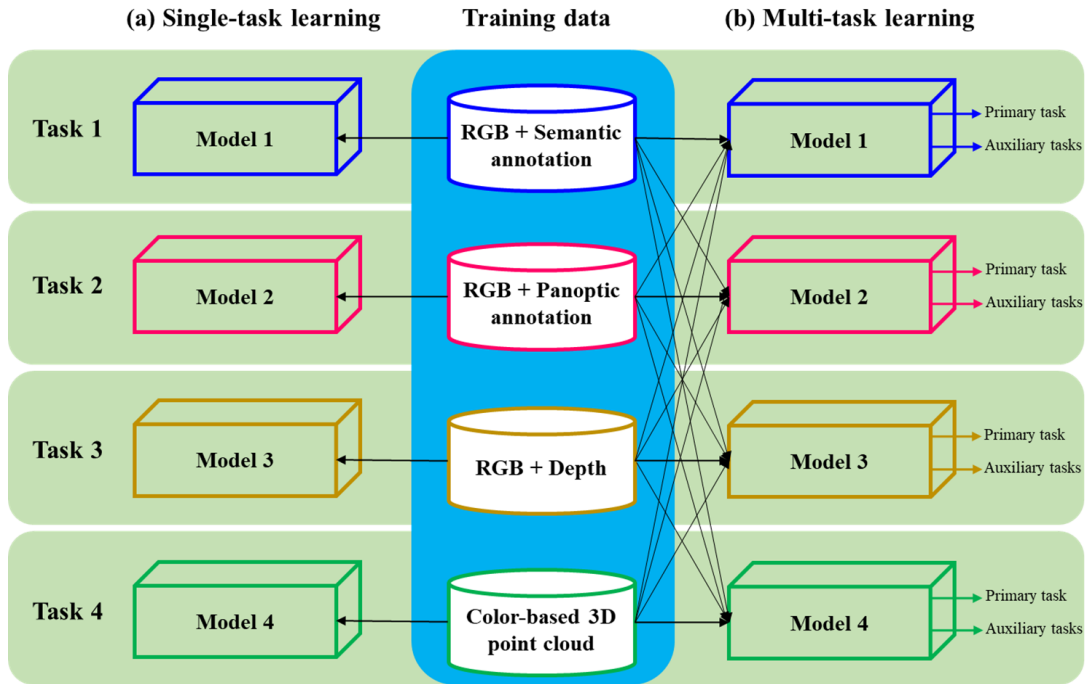

Figure 1. Difference between (a) single-task learning, and (b) multi-task learning

## 4. RESULTS

This study aims to promote and strengthen the research on multi-task learning for scene understanding from fixed-wing UAV images. The proposed method is composed of two components of UAV-based multi-task data generation and evaluation methods in urban scene analysis. The contributions of this study are as follows.
(1) Development of a multi-task dataset for scene understanding, including very high-resolution RGB orthophoto, digital surface model, semantic annotation, building panoptic annotation (cf. Figure 3).
(2) An evaluation is presented based on the most important challenges for multi-task data analysis for scene understanding.
In this study, we will build a large multi-task dataset for urban scene analysis. The study area is located in Esfahan, Iran. The study region includes buildings with rectangular flat roofs which may also have various tiny structures. Flat roof building is a specific style of urban architecture common in different cities (e.g., New York, Cairo, Tehran, and Esfahan). The planned specifics of the dataset are listed in Table 2. Moreover, Figure 4 shows the study site in the various landscape types with three samples of datasets collected.

| GSD | Image size | Tile | Semantic classes | | Panoptic class | DSM |
|---|---|---|---|---|---|---|
| 2.5cm | 2000×2000pix | 500 | (1) Parcel boundary<br>(2) Vehicle<br>(3) Building shadow<br>(4) Vegetation<br>(5) Building | (6) Ground/Road<br>(7) Waterbodies<br>(8) Farmland<br>(9) Waste object<br>(10) Lane-marking | Building | Yes |

Table 2. Detail of the proposed multi-task dataset

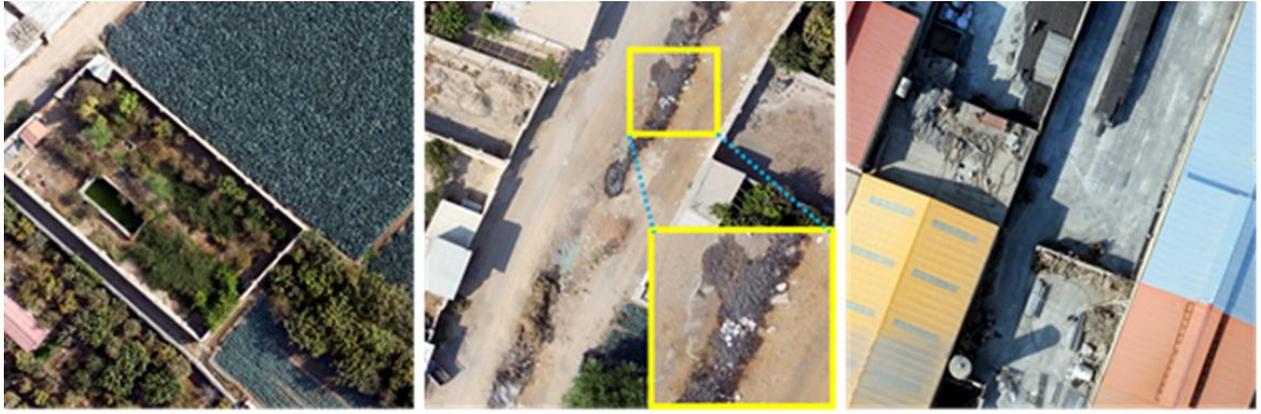

Figure 4. Various landscape types in the proposed dataset: (a) water and vegetation covers, (b) waste objects, and (c) different color roofs

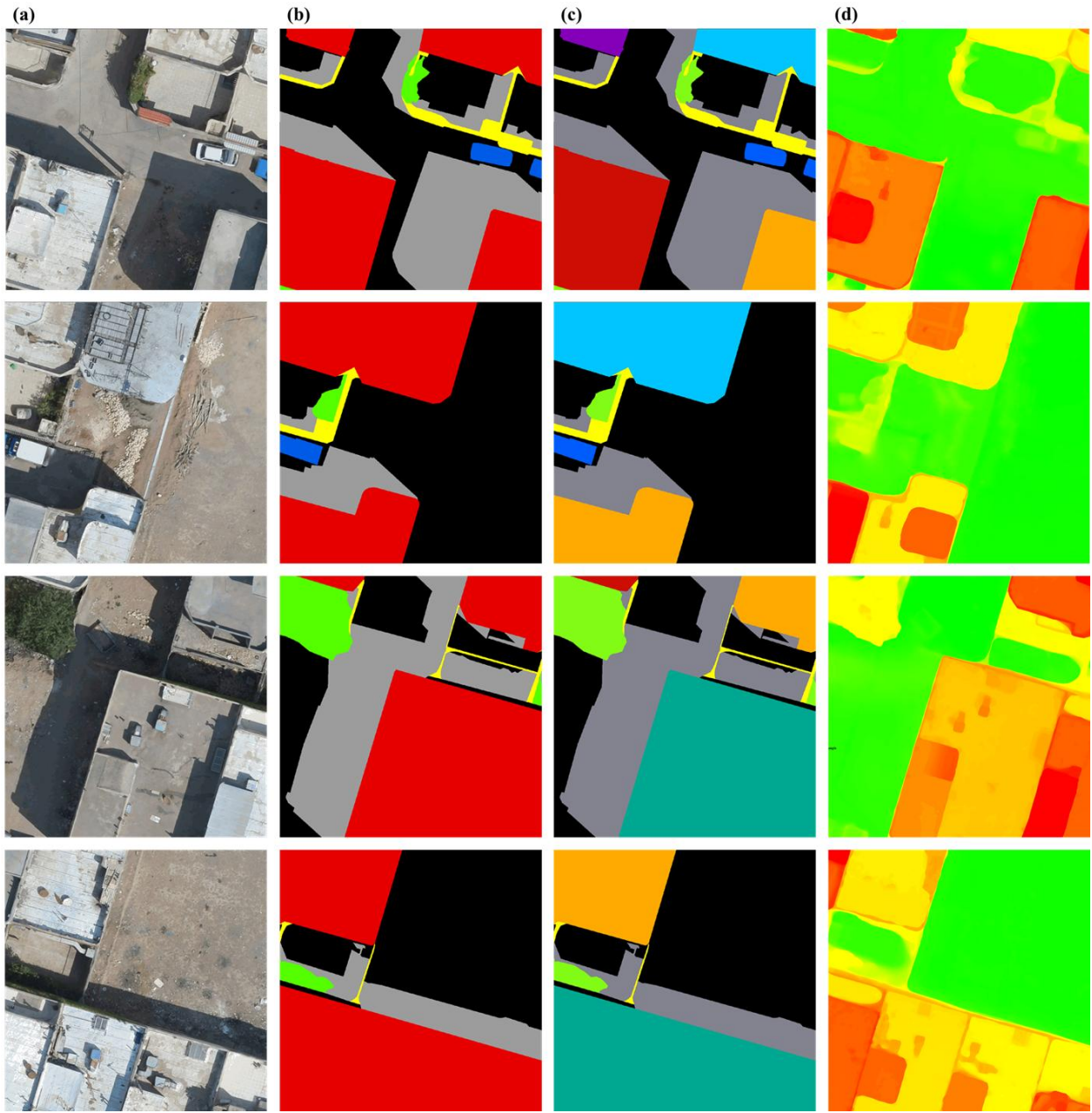

Figure 3. Sample screenshot of the proposed dataset for scene understanding with (a) RGB orthophoto, (b) semantic segmentation annotation, (c) building panoptic annotation, and (d) DSM

## 5. CONCLUSION

In this study, a common framework and a new dataset for the performance assessment of multi-task learning methods from fixed-wing UAV images for 2D/3D city modeling are presented. The UAV RGB images were captured by SAMA-VTOL are provided for the case study. This fixed-wing UAV has an installed very high-resolution Fujifilm sensor that was intended for professional photogrammetry. This UAV can cover 100 ha during one flight and the flight time is 60 min. The results of the experiments in the test area indicate that the SAMA-VTOL is robust to multi-task data generation.